\documentclass[10pt,twocolumn,letterpaper]{article}

\usepackage{cvpr}
\usepackage{times}
\usepackage{epsfig}
\usepackage{graphicx}
\usepackage{amsmath}
\usepackage{amssymb}

\usepackage[pagebackref=true,breaklinks=true,letterpaper=true,colorlinks,bookmarks=false]{hyperref}

\cvprfinalcopy 

\def\cvprPaperID{6749} 

\begin{document}

\title{Direct Object Recognition Without Line-of-Sight Using Optical Coherence}



\author{Xin Lei\textsuperscript{1}$^*$ \ \  Liangyu He\textsuperscript{1}$^*$ \ \ Yixuan Tan\textsuperscript{1}$^*$ \ \ Ken Xingze Wang\textsuperscript{1}$^{\dag}$ \ \  Xinggang Wang\textsuperscript{2} \ \    Yihan Du\textsuperscript{2}\\  Shanhui Fan\textsuperscript{3} \ \  Zongfu Yu\textsuperscript{4}\\
        \textsuperscript{1}Coherent AI LLC\\
        \textsuperscript{2}Institute of AI, School of EIC, Huazhong University of Science and Technology\\
        \textsuperscript{3}Ginzton Laboratory, Department of Electrical Engineering, Stanford University\\
        \textsuperscript{4}Department of Electrical and Computer Engineering, University of Wisconsin-Madison\\
    \tt\small $^{\dag}$ken@coherent.ai
}

\maketitle

\begin{abstract}
Visual object recognition under situations in which the direct line-of-sight is blocked, such as when it is occluded around the corner, is of practical importance in a wide range of applications. With coherent illumination, the light scattered from diffusive walls forms speckle patterns that contain information of the hidden object. It is possible to realize non-line-of-sight (NLOS) recognition with these speckle patterns. We introduce a novel approach based on speckle pattern recognition with deep neural network, which is simpler and more robust than other NLOS recognition methods. Simulations and experiments are performed to verify the feasibility and performance of this approach.
\end{abstract}


\let\thefootnote\relax\footnote{$^*$Equal contribution $^{\dag}$Corresponding author}

\section{Introduction}

Object recognition is essential for various applications such as face recognition, industrial inspection, medical imaging, and autonomous driving. One intriguing area of research is the recognition of objects without direct line-of-sight~\cite{Adib,Balaji,Bouman,Heide,Katz,Kirmani2,Kirmani,OToole,Pandharkar,Velten}. The ability to carry out recognition without line-of-sight has practical significance; for example, in autonomous driving, if the imaging system can recognize pedestrians and vehicles that are ``hidden'' around the corner or behind other obstacles, the vehicle can prevent potential hazards and greatly increase the safety level of driving.

\begin{figure}[hbtp]
\begin{center}
   \includegraphics[width=0.9\linewidth]{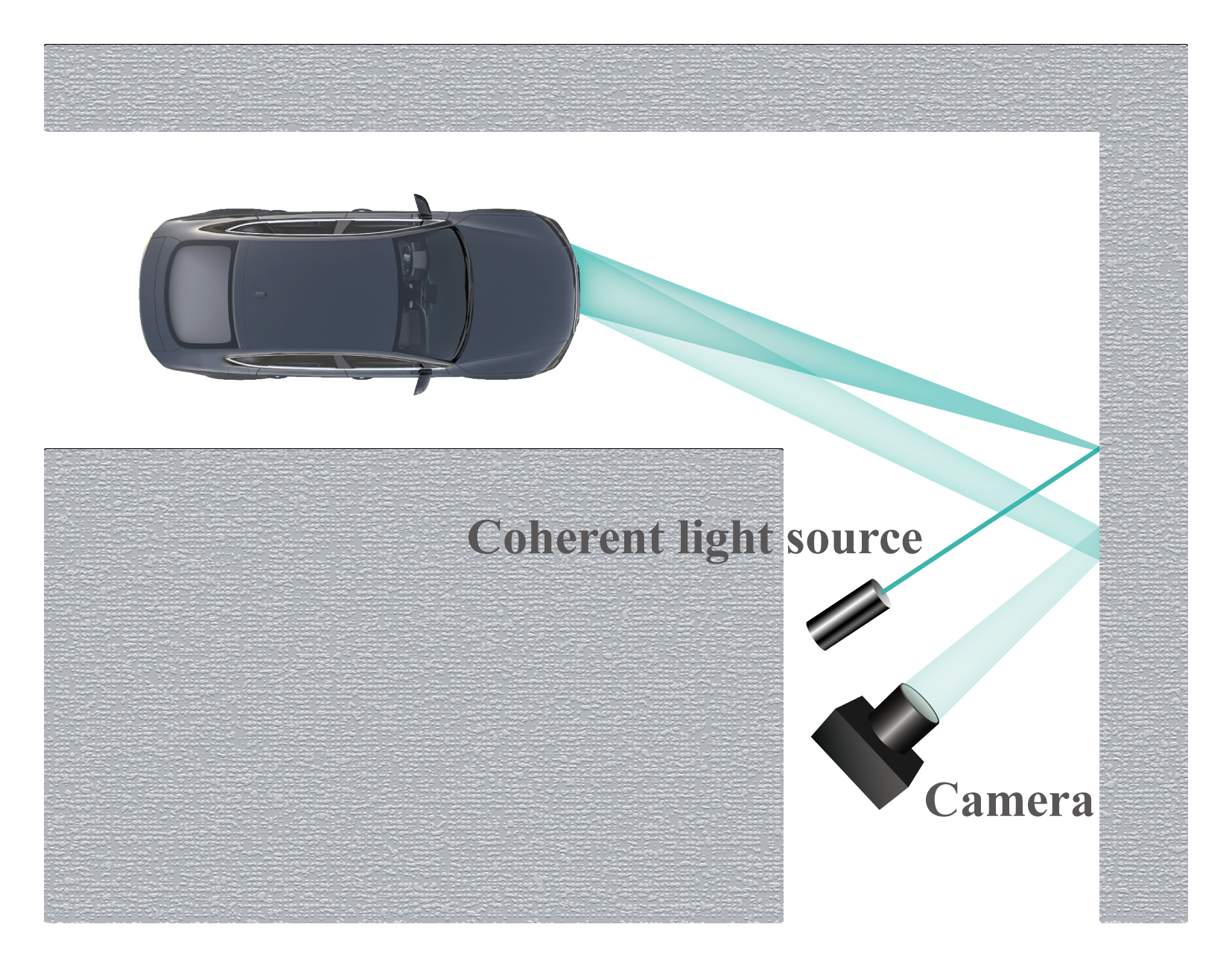}
\end{center}
   \caption{Object recognition without direct line-of-sight. Coherent light source such as laser is used to illuminate the hidden object, which is a car in this example. Light scattered by the object forms speckle patterns on the wall which is captured by the camera for recognition.}
\label{fig:intro}
\end{figure}

In this paper, we report an approach to perform recognition without direct line-of-sight by exploiting the coherent light. Objects illuminated by coherent light sources such as lasers form speckle patterns due to interference. In a non-line-of-sight situation, for example, as shown in Figure~\ref{fig:intro}, the object itself is occluded from the camera; however, the speckle patterns on the wall are a result of the interference of light scattered owing to the object~\cite{Ando,Horisaki,Katz2,Lyu,Satat,Shih,Sinha,Smith,Valent,Yoo}. The information in these patterns is not obvious to a human, but with an appropriate deep network, one can retrieve the information necessary to identify the object. Compared to current computer vision algorithms that are primarily based on direct visual images, this approach is applicable to a wide range of situations in which visual images of the objects cannot be obtained. 

Our primary contribution is a novel NLOS object recognition method that uses the information contained in speckle patterns under coherent illumination. We use this method to realize direct recognition of object without line-of-sight by employing inexpensive electronic devices. In addition to the simplest situation of scattering from one diffusive wall, we demonstrate the feasibility of this method in several more challenging scenarios: (1) when the light source and the camera locate at the same side so that neither of them has direct line-of-sight with the object; this allows for wider application of our method in which the object space is not accessible; (2) when the wall is randomly rotated after each measurement; this shows that our method is independent of the specific textures of the wall and is robust; (3) when there are two scatterings from two diffusive walls before the light reaches the camera; this shows the potential of our method to allow multiple scatterings in a complex scene.

We quantitatively evaluate the performance of our preliminary setups using both experiments and simulations on hand-written digits (MNIST) as well as much more complex visual objects (human body posture) in the aforementioned scenarios. We demonstrate recognition accuracy of over 90\% with MNIST dataset under various experimental and simulation conditions. For human body posture dataset, an accuracy of 78.18\% is achieved for classification of 10 body posture categories, higher than the three-way classification result of body posture reported in~\cite{Satat2}.

\section{Related Work}

Recently, a few interesting works have demonstrated the formation of images of objects that are without direct line-of-sight. These methods overcome the traditional limitations of imaging optics, which cannot form clear images in the absence of direct line-of-sight conditions. These experiments are usually extensive, requiring non-traditional measurement of light, for example, by measuring time-of-flight (TOF) or in a setup that preserves the memory effect. However, imaging is not always needed for recognition. To perform imaging without line-of-sight, one would require expensive hardware and suffer from practical limitations such as a narrow field-of-view. In this study, we perform direct recognition without imaging the object. This aspect allows our method to overcome some of the critical practical limitations of the related imaging methods. The proposed method requires hardware (mostly consumer-grade electronics) that is far less expensive than that required for the TOF-based NLOS imaging~\cite{Kadambi,Marco,Naik2,Naik,OToole,Shrestha,Su,Velten,Wu}, and the method is more robust than the memory-effect based imaging techniques that have a limited field-of-view~\cite{Guo,Li}. Moreover, a recent publication~\cite{OToole18} also uses only ordinal digital cameras but would require very specific scene setup (an accidental occlusion) to obtain better performance. 

\subsection{Imaging Based on Time-of-Flight}

NLOS imaging with TOF has recently received considerable attention. It is a range imaging system that resolves the distance based on measuring the TOF of a light signal between the object and the camera for each point of the image. The mechanism of TOF measurement without line-of-sight is as follows~\cite{Velten}: A laser pulse hits a wall that scatters the light diffusely to a hidden object; then, the light returns to the wall and is captured by a camera. By changing the position of the laser beam on the wall with a set of galvanometer-actuated mirrors, the shape of the hidden object can be determined. 

Although TOF measurements without line-of-sight can recover the hidden object with an accuracy of the order of centimeters, it has some disadvantages. For example, it requires substantial resources such as single-photon detectors and nanosecond-pulsed lasers, which could cost tens of thousands of dollars. In contrast, our method only uses standard lasers and CMOS image sensors, which cost only some thousands of dollars. TOF also takes minutes for data acquisition and image reconstruction, while our approach is only a single-shot measurement that takes less than one second. 

\subsection{Imaging Based on Speckle Correlation (Memory Effect)}

Imaging via speckle correlation is another method that was recently developed. When a rough surface is illuminated by a coherent light (\eg, a laser beam), a speckle pattern is observed in the image plane. The key principle of this method is that the auto-correlation of the speckle pattern is essentially identical to the original object's auto-correlation, as if it is imaged by a perfect diffraction-limited optical system that has replaced the scattering medium. Consequently, the object's image can be obtained from its auto-correlation by an iterative phase retrieval algorithm~\cite{Katz}. In particular, for seeing without line-of-sight, the light back-scattered from a diffusive wall is used to image the hidden objects. 

For this method to work, the auto-correlation must be preserved, which limits the field-of-view. To use auto-correlation to recover an image, the image must be very sparse. Compared to this imaging technique, our approach is more robust. It does not depend on the scattering property of the wall and a single neural network can be trained to work for many different types of walls. It also does not require the object to be sparse, and there is no limit on the field-of-view. 

\subsection{Imaging Based on Holographic Approach}

In 2014, Singh \etal proposed a holographic approach for visualizing objects without line-of-sight, based on the numerical reconstruction of 3D objects by digital holography in which a hologram is formed on a reflectively scattering surface~\cite{Singh}. A coherent light source is divided into two parts: One beam illuminates the object, while the other is set as the reference beam. The interference between the two beams forms an aerial hologram immediately in front of the scattering surface. Then, the hologram is recorded by a remote digital camera that focuses on the scattering surface. 

This holographic technique requires a reference beam for holographic recording, which is particularly challenging to implement in practical scenarios.

\section{Method}

\subsection{Preliminary Knowledge}

Light carries information in terms of not only intensity but also phase, frequency, and polarization, among other factors. Traditional imaging techniques utilize only the intensity information by using photon-electron conversion through active semiconductor materials. Ordinary light sources emit incoherent light; therefore, the phase of each wave packet is random, and it cannot be used to retrieve scene information. Considering an object with two points reflecting incoherent light, the total intensity $I$ at the image plane is the sum of the intensity of the two point sources $I_1$ and $I_2$, and the phase information is effectively lost:

\begin{equation}
I=I_1+I_2.
\label{eq:inc_2p}
\end{equation}

The light emitted by coherent light sources such as lasers can form interference patterns. The total intensity with coherent illumination contains a phase term, which is a function of the phase difference $\Delta\phi$; it is highly dependent on the relative locations of the two point sources, and results in interference patterns~\cite{Goodman}:

\begin{equation}
I=I_1+I_2+2\sqrt{I_1I_2}\cdot\cos{\Delta\phi}.
\label{eq:co_2p}
\end{equation}

An object illuminated by either incoherent or coherent light can be considered as a secondary light source. For incoherent light, a pixel at location $\vec{r}$ on the image plane captures light from all over the object:

\begin{equation}
I(\vec{r})=\int\left|E(\vec{R},\vec{r})\right|^2\cdot d\vec{R},
\label{eq:inc_int}
\end{equation}
where $E(\vec{R},\vec{r})$ is the amplitude distribution at the image plane from a point source located at $\vec{R}$ on the object. The phase variation of different pathways does not affect the intensity. Traditional imaging approaches using incoherent light therefore do not utilize the phase information of light. Without direct line-of-sight to perform imaging, the object can hardly be recognized under incoherent illumination. 

For coherent light, the light intensity is given by

\begin{equation}
I(\vec{r})=\left|\operatorname{Re}\left[\int E(\vec{R},\vec{r}) \cdot \operatorname{exp}(i\phi(\vec{R},\vec{r})) \cdot d\vec{R}\right]\right|^2,
\label{eq:co_int}
\end{equation}
where $\phi(\vec{R},\vec{r})$ is the phase variation at the image plane from a point source located at $\vec{R}$ on the object, which is highly dependent on the object geometry and its location with respect to the image plane. Objects with a complex surface geometry at the wavelength scale therefore generate complex, seemingly random interference patterns $I(\vec{r})$ --- more specifically, speckle patterns, a term commonly used in laser studies. The phase information is represented by the bright and dark distributions of light intensity in the speckle pattern. Therefore, the speckle pattern $I(\vec{r})$ contains information of the object, and an appropriate deep learning network can effectively use such information to perform object recognition.

\begin{figure}[hbtp]
\begin{center}
   \includegraphics[width=1\linewidth]{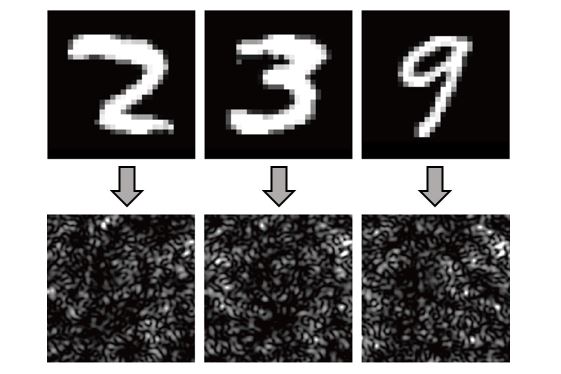}
\end{center}
   \caption{Speckle patterns of hand-written digits from simulation. The images in the top row are original images of hand-written digits from the MNIST dataset, and the images in the bottom row are speckle patterns corresponding to the digits in the top row.}
\label{fig:prelim}
\end{figure}

Consider the example of the MNIST hand-written digits. Speckle images of several different hand-written digits are shown in Figure~\ref{fig:prelim}. As described in Equation~\ref{eq:co_int}, each point on a hand-written digit forms a secondary point source under coherent illumination, whose responses at the image plane are integrated to form the speckle patterns shown in the bottom row of Figure~\ref{fig:prelim}. Those speckle patterns are intensity distributions that can be captured by traditional image sensors. After training with tens of thousands of speckles from different hand-written digits, the deep network is able to find meaningful invariant features among speckles generated from the same digit with various hand-written styles, and infer the properties of these objects that have interacted with the light during its propagation. In other words, with well-trained deep networks, one can retrieve the information from speckle patterns to identify the object.

\subsection{Simulation and Experiment Methods}

\begin{figure}[hbtp]
\begin{center}
   \includegraphics[width=1\linewidth]{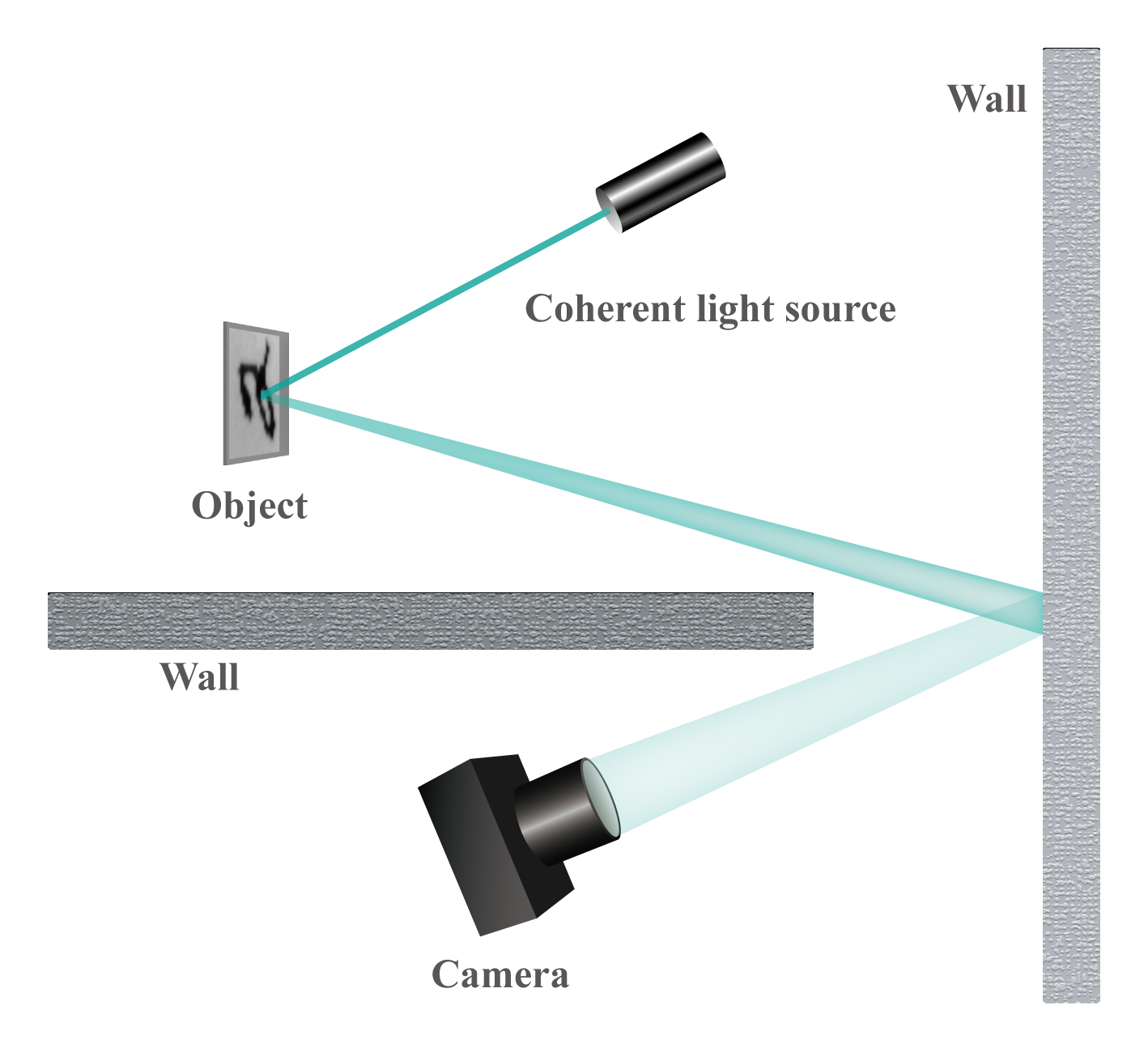}
\end{center}
   \caption{Schematic of object recognition with one wall as the scattering surface. The wall on the left side blocks the direct line-of-sight between the object and the camera.}
\label{fig:one_wall}
\end{figure}

We supported the above theory by performing a set of experiments and simulations. To recognize objects without line-of-sight, the simplest situation is shown in Figure~\ref{fig:one_wall}. The direct line-of-sight between the object and the camera is blocked by a wall; therefore, direct images cannot be formed through traditional imaging systems. However, in many situations, it is not difficult to find another surface (usually a diffusive surface) that can reflect part of the light from the object to the camera. A typical example is another wall, which is located at a convenient place, as depicted in Figure~\ref{fig:one_wall}. Under coherent illumination, the light scattered off the object forms speckle patterns on the ``relay'' wall, which is captured by the camera and understood by the deep network. The coherent illumination covers the object to be detected.

Both the simulation and the experiment were performed under this scheme. In the simulation, the objects were modeled as opaque boards with the transparent features of the digits allowing light to pass. Coherent illumination was modeled as a simple plane wave. The Fourier optics method was used to simulate the light propagation process. The wall was modeled as a scattering object with pixelated random phase modulation. The lens was modeled as a phase modulation element with perfect image forming ability. 

\begin{figure}[hbtp]
\begin{center}
   \includegraphics[width=0.95\linewidth]{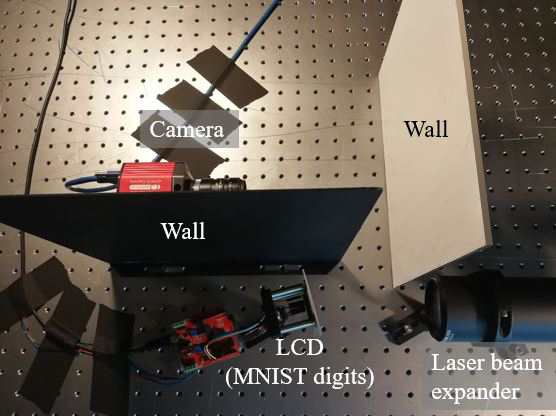}
\end{center}
   \caption{Experimental setup for object recognition with one wall.}
\label{fig:one_wall_exp}
\end{figure}

In the experiment, as shown in Figure~\ref{fig:one_wall_exp}, the object was a reflective LCD screen (2.5 cm $\times$ 2.5 cm) displaying the images of hand-written digits from the MNIST dataset. The camera was an off-the-shelf camera (Thorlabs CS2100M CMOS) with a lens (Thorlabs MVL35M23) focused on the wall, which is an aluminum board (30.5 cm $\times$ 30.5 cm) painted with white egg-shell wall paint to emulate a real indoor diffusive wall surface. A HeNe laser (632.8 nm, Thorlabs HNL150R) was used to provide coherent illumination to the object, which covered the whole LCD screen. Another wall was installed between the camera and the LCD to completely block the direct line-of-sight between them. The distances between the object, camera and the wall are approximately 20 cm.

\begin{figure}[hbtp]
\begin{center}
   \includegraphics[width=1\linewidth]{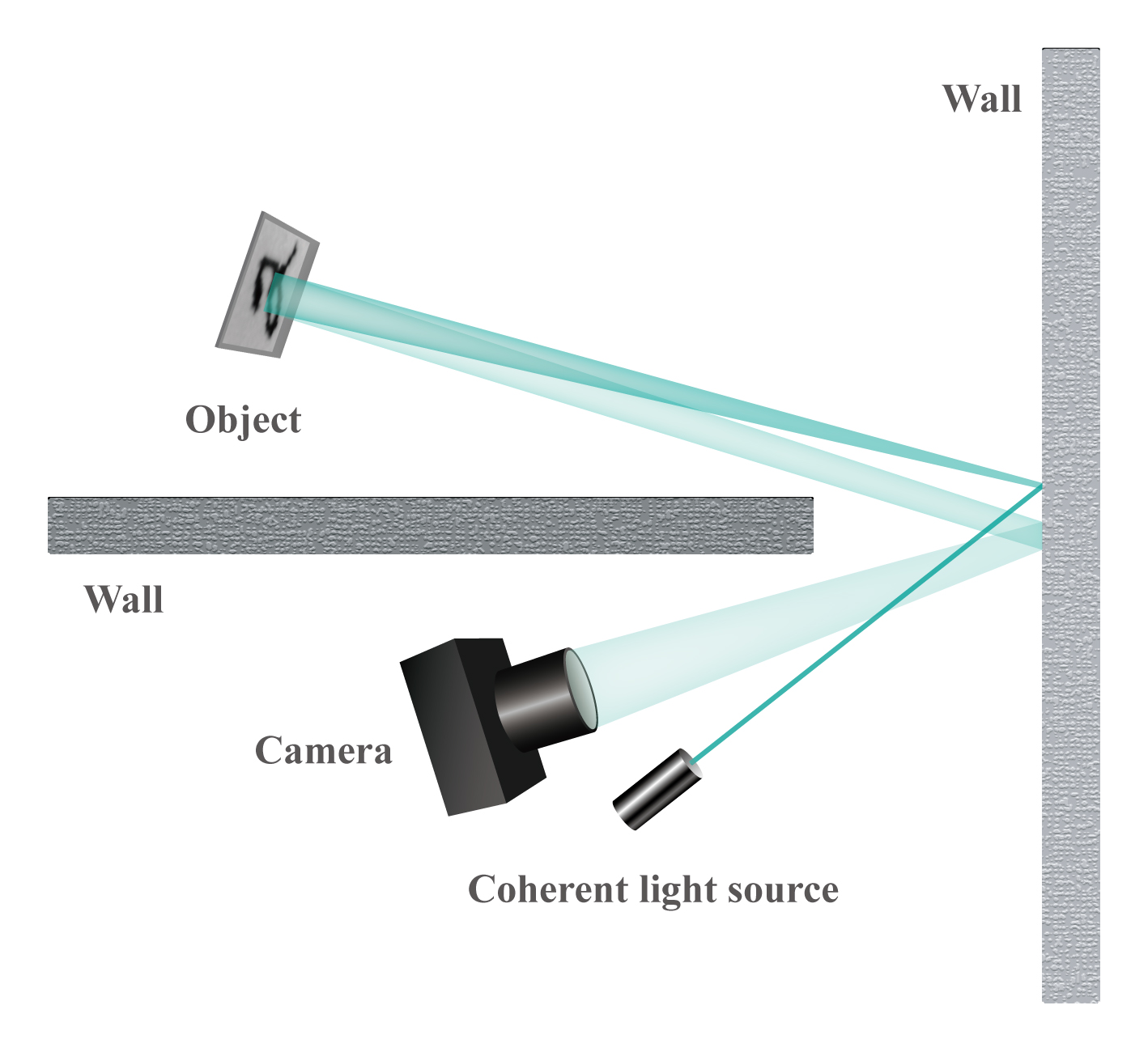}
\end{center}
   \caption{Schematic of object recognition with the coherent light source on the same side as the camera.}
\label{fig:same_side}
\end{figure}

In real application, however, it is often impossible to directly access the occluded object to provide coherent illumination. One alternative way is to use the ``relay'' wall to reflect the coherent light into the ``object'' space, as shown in Figure~\ref{fig:same_side}. This scheme allows for much wider application of this technology. To verify its feasibility, we performed the simulation as follows. The light source was modeled as a point source that propagates to the wall; the light is scattered, and then it propagates to the object. The subsequent steps are similar to the simulation process described above. In practice, the patch of wall surface that the camera is pointed at should be away from the laser spot on the wall to avoid potential saturation of the sensor from the high intensity laser beam.

\begin{figure}[hbtp]
\begin{center}
   \includegraphics[width=1\linewidth]{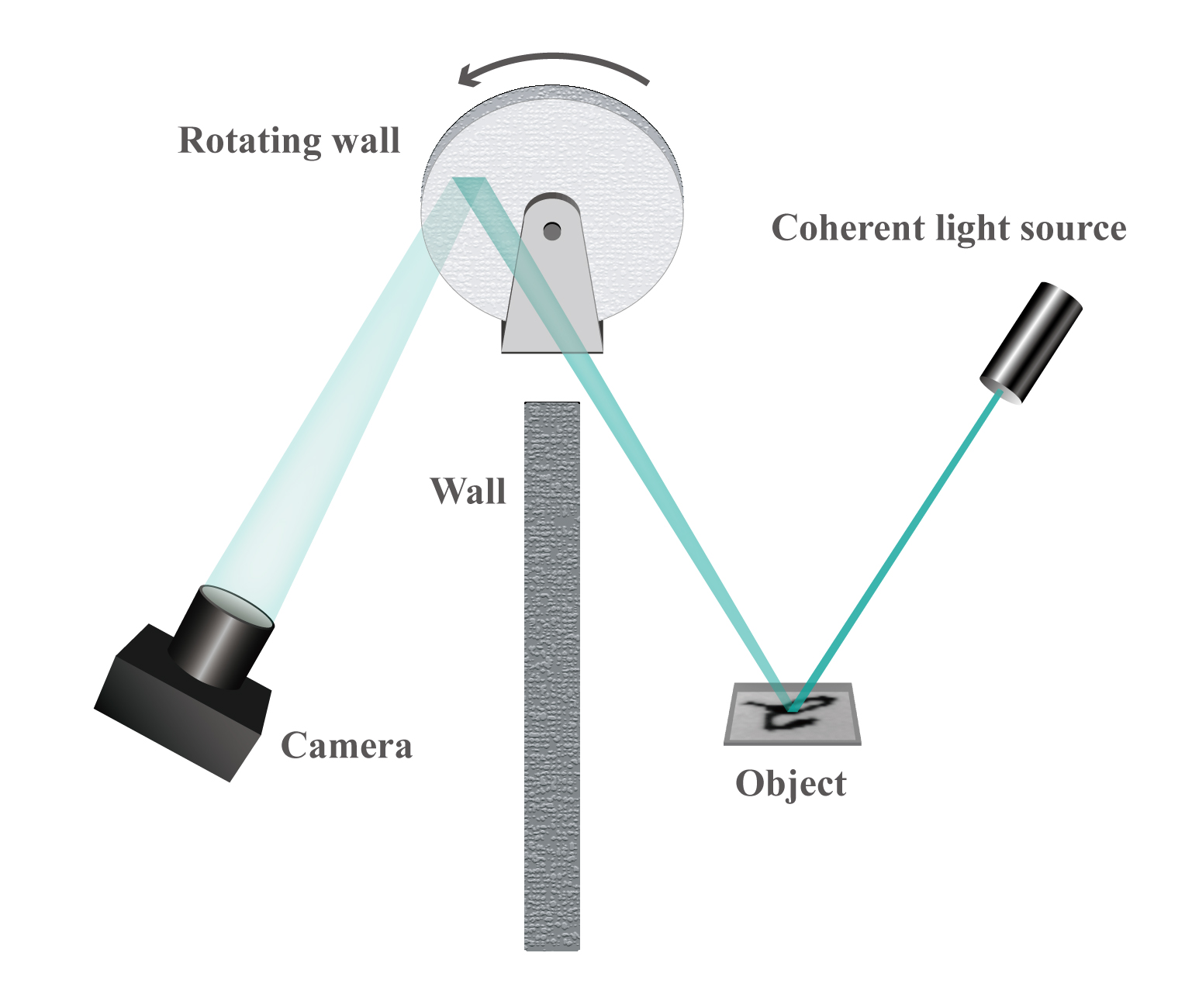}
\end{center}
   \caption{Schematic of object recognition with a rotating wall.}
\label{fig:rotate_wall}
\end{figure}

Theoretically, the speckle pattern on the wall is determined to the first order by the object scattering the light, and it should stay relatively invariant within a range of varying wall surface conditions. The surface condition might modulate the speckle patterns and appear as increased noise. To verify the feasibility, we designed an experiment in which the wall is rotating, as shown in Figure~\ref{fig:rotate_wall}. In the experiment, the wall was a round plastic plate painted with white egg-shell wall paint. The camera lens was focused on the outer rim of the plastic plate. We rotated the wall by a random degree after every capture so that each instance of the object was effectively reflected by a different random patch of the wall. This experiment tested the robustness of this scheme.

\begin{figure}[hbtp]
\begin{center}
   \includegraphics[width=1\linewidth]{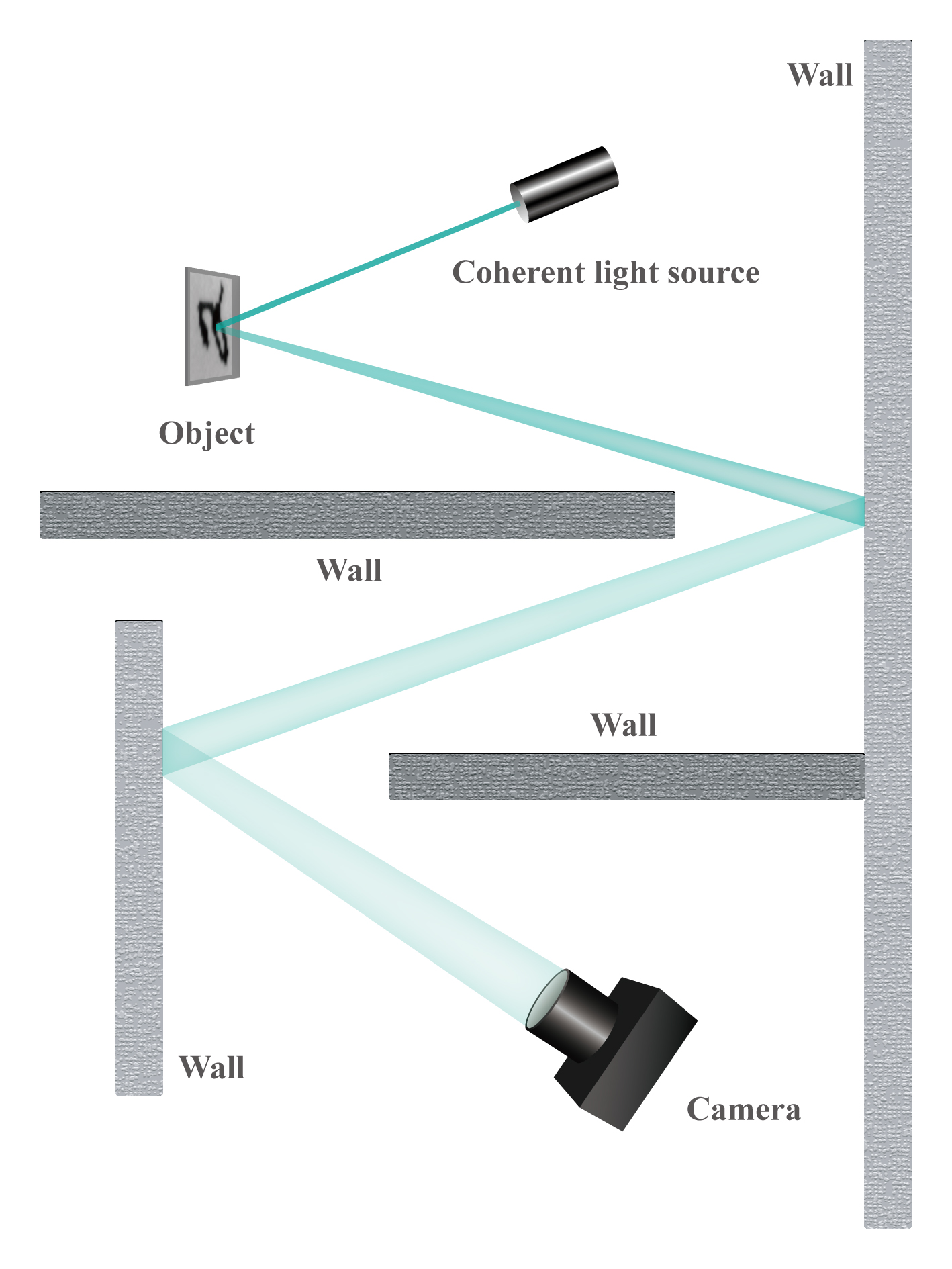}
\end{center}
   \caption{Schematic of object recognition with two walls.}
\label{fig:two_wall}
\end{figure}

We further extended the capability of this scheme to situations with two diffusive walls, as shown in Figure~\ref{fig:two_wall}. In the simulation, the second wall was added after the first wall, and the camera lens focused on the second wall that was directly visible to the camera. This opened up the possibility to use this technology in much more complex environments in which the light from the object undergoes multiple reflections before reaching the camera.

\subsection{Data Preprocessing}

Data (speckle patterns) from both the simulation and the experiment need to be preprocessed before training with the deep network. (1) Due to computational limitations, we only cropped a small part of the output speckle images. This process reduces both the number of network parameters and the required data size for training without overfitting~\cite{Li2}. Because each part of the output speckle image contains information of the whole object, even a small part of the speckle image can be used for object recognition. In other words, image cropping does not affect the final recognition results. (2) In a real environment for recognition, there could be various disturbances arising from temperature variation, vibration, drifting, etc. Therefore, adding simulated random noises to the speckle patterns could increase the recognition accuracy. (3) The speckle images were normalized between 0 and 1. Typically, two datasets were used in this study: MNIST dataset of hand-written digits and human body posture dataset. 

In the experiment, because the LCD screen had only the on and off states, the images of the hand-written digits from the MNIST dataset were binarized for display on the LCD screen. The first 10,000 hand-written digits in the MNIST were used in both the experiments and the simulations.

The posture dataset was preprocessed before the simulation by using DeepLab-v3~\cite{Chen2,Chen} as a human detector to remove the background. It was also converted to the gray scale. 

\subsection{Deep Networks for Speckle Pattern Recognition}
\label{sec:net}

\begin{figure*}[hbtp]
\begin{center}
   \includegraphics[width=1\linewidth]{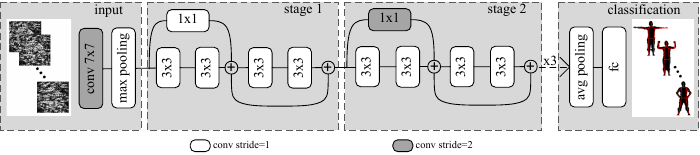}
\end{center}
   \caption{Human body posture classification using ResNet-18.}
\label{fig:resnet}
\end{figure*}

We used deep networks to dig the semantic information in the speckle pattern images to relate them to the original objects, though the images were not recognizable by humans. In our experiments and simulations, the object recognition problems were typical image classification problems. Thus, we could follow the common practices in computer vision to train and test the classification networks. We applied two networks, a simple network (termed SimpleNet) and ResNet-18~\cite{Cao,Geifman,He}. 

SimpleNet contains 4 convolutional layers with ReLu activation. The first two convolutional layers have 16 3$\times$3 filters (with 2$\times$2 max pooling), and the third and fourth convolutional layers have 32 3$\times$3 filters (with 2$\times$2 max pooling). A fully connected layer with 1024 neurons is used right before the output layer of 10 neurons for 10 digits. SimpleNet has high training and testing speed and it was used in our MNIST experiments. The speckle patterns were cropped to 200$\times$200 and fed into SimpleNet. Among the 10,000 samples, a randomly chosen subset of 95\% of the samples were used for training and the remaining 5\% were used for testing.

ResNet is widely applied in understanding visual images. Due to its unique residual connection design shown in Figure~\ref{fig:resnet}, it is easy to train very deep residual networks. We use the ImageNet pretrained ResNet models which are publicly available in PyTorch~\cite{pytorch}. The last full connection layer was replaced with 10 neurons for 10 class human body posture classifications. An input speckle pattern image was first resized to 256$\times$256; then randomly cropped 224$\times$224 patches from it were used in training. For testing, a speckle pattern image was directly resized to 224$\times$224 and then fed into ResNet. We chose ResNet-18 for its high-speed, high-performance and good generalization ability.

\section{Experiments}

\subsection{Experimental Results for MNIST}

\begin{figure}[hbtp]
\begin{center}
   \includegraphics[width=1\linewidth]{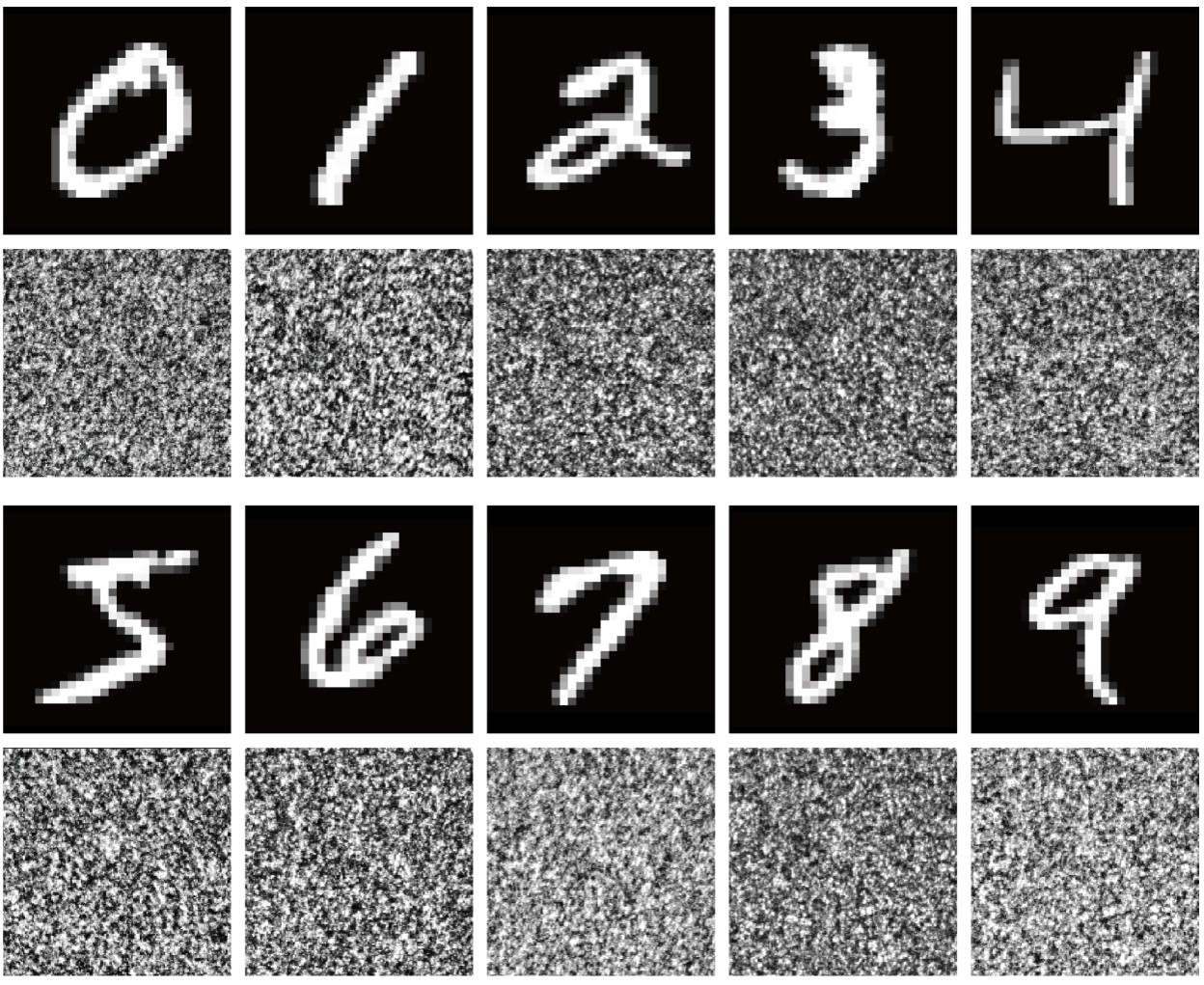}
\end{center}
   \caption{Speckle patterns of different hand-written digits captured on the CMOS image sensor in the experiment.}
\label{fig:speckle_exp}
\end{figure}

We first used the MNIST dataset to demonstrate its feasibility with the simplest situation where the light scattered off an object reflects only once before reaching the camera, as shown in Figure~\ref{fig:one_wall}. Both an experiment and a simulation were performed. Examples of speckle images on the wall from the experiment are shown in Figure~\ref{fig:speckle_exp}. All images are full of speckles, indicating strong scattering and interference effects. The deep network used in the experiment is SimpleNet defined in Section~\ref{sec:net}. Despite the lack of a direct line-of-sight, the recognition accuracy in the experiment and the simulation are 95\% and 97\% respectively. As expected, the experiment has slightly lower accuracy, likely due to noise, of which there were multiple sources, including image sensor noise, ambient light, temperature fluctuation, vibration, and movement of each element used in the experiment which could cause speckle pattern drifting.

To demonstrate the feasibility of having a laser and a camera on the same side of the blocking wall, as shown in Figure~\ref{fig:same_side}, we performed a simulation using the MNIST data. In this case neither the light source nor the detector had a direct line-of-sight of the object. This arrangement is much more widely applicable than the previous situation. Simulation shows that the recognition accuracy is 97\%, which is not very different from that in the previous case. In the experimental demonstration, the light intensity of the laser will be greatly reduced after it diffusively reflects from the wall. Further, the signal-to-noise ratio could be much smaller and we expect the recognition accuracy to decrease. It would, however, require additional engineering effort to achieve satisfying performance. Nonetheless, this simulation results demonstrate the possibility of having the laser and detector on the same side, which opens up opportunities for a wide range of applications.

To demonstrate the robustness of our method, we performed an experiment where the wall is rotated by a random degree after each measurement and every example of the object is effectively reflected from a different patch on the wall, as shown in Figure~\ref{fig:rotate_wall}. Recognition accuracy is 91\%, slightly lower than that for a still wall, but still significantly high. It shows that this configuration of NLOS recognition is independent of the specific textures of the wall.

\begin{table*}
\begin{center}
\begin{tabular}{|l|c|}
\hline
Experiments & Recognition Accuracy \\
\hline\hline
One wall (simulation) & 0.97 \\
One wall (experiment) & 0.95 \\
One rotating wall (experiment) & 0.91 \\
One wall, laser and camera on the same side (simulation) & 0.97 \\
Two walls (simulation) & 0.97 \\
\hline
\end{tabular}
\end{center}
\caption{Deep network recognition accuracy when using MNIST data. Recognition accuracy from experiments and simulations are well above 90\%.}
\label{tab:mnist}
\end{table*}

Furthermore, we extended the simulation to add another diffusive wall to the setup, as shown in Figure~\ref{fig:two_wall}. This increases the complexity of the scene. The recognition accuracy is 97\%, similar to the situation of having only one wall. The simulation results demonstrate the power of this scheme, which potentially allows object recognition after multiple reflections in a complex scene.

A summary of the recognition results of the experiments and the simulations using the MNIST dataset is shown in Table~\ref{tab:mnist}.

\subsection{Experimental Results for Human Body Posture}

\begin{figure}[hbtp]
\begin{center}
   \includegraphics[width=1\linewidth]{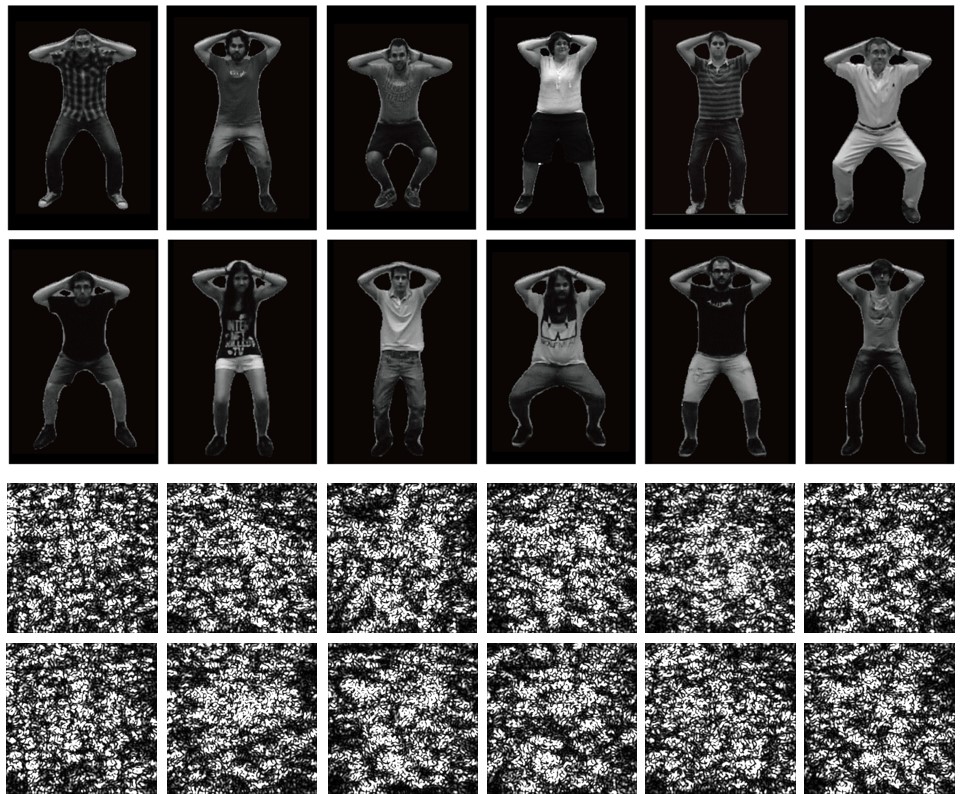}
\end{center}
   \caption{Images from the human body posture dataset showing the same posture of the 12 different human subjects. The related speckle patterns obtained via simulation are shown in the bottom rows.}
\label{fig:posture}
\end{figure}

The MNIST dataset is relatively simple to classify. Complex objects with more detailed features and greater variations in the shape could result in more challenges to the recognition task. Here, we used human body posture as an example to demonstrate the recognition of complex objects. We performed a simulation following the scheme shown in Figure~\ref{fig:one_wall} with one still wall and using the human body posture dataset. The dataset contains 10 posture categories, and the image data is collected from 12 human subjects~\cite{bodypose}. Images in this dataset contain much more complex features of the human body than the hand-written digits, and the same body posture shows appreciable variations among different human subjects, as shown in Figure~\ref{fig:posture}. The recognition task was therefore much harder compared to the situation using the MNIST dataset. We extracted the human body posture data using a DeepLab-v3 person segmentation model~\cite{Chen2,Chen} to remove the background, based on the assumption that in real applications the background is often at a distance from the object to be recognized and therefore, light scattered from the object will have a larger impact on the speckle pattern on the wall. We also converted the color images to the gray scale because the coherent source is monochromatic.

\begin{figure}[hbtp]
\begin{center}
   \includegraphics[width=1\linewidth]{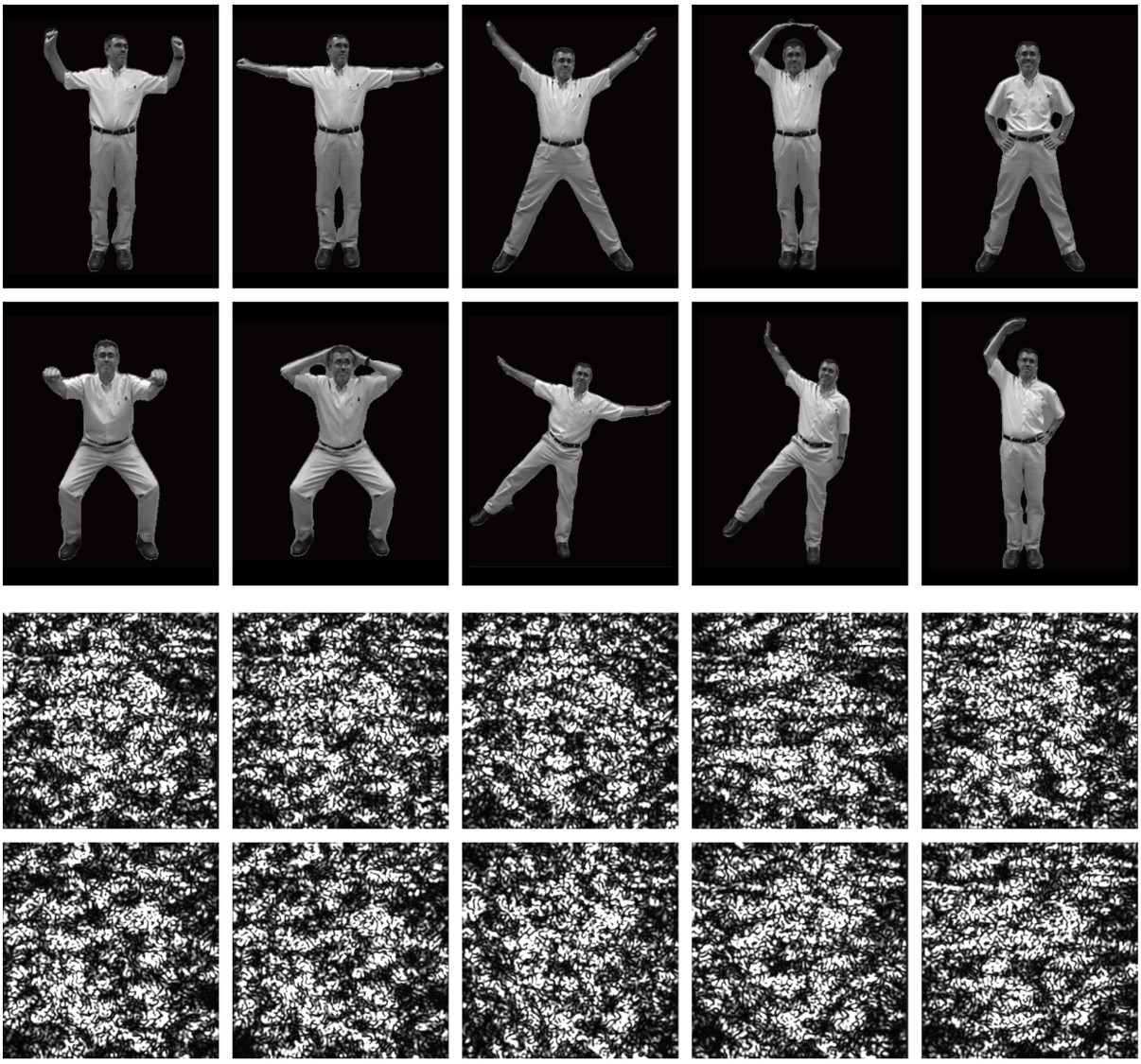}
\end{center}
   \caption{Images from the human body posture dataset showing the 10 different postures of the same human subject. The related speckle patterns obtained via simulation are shown in the bottom rows.}
\label{fig:posture_speckle}
\end{figure}

\begin{figure}[hbtp]
\begin{center}
   \includegraphics[width=1\linewidth]{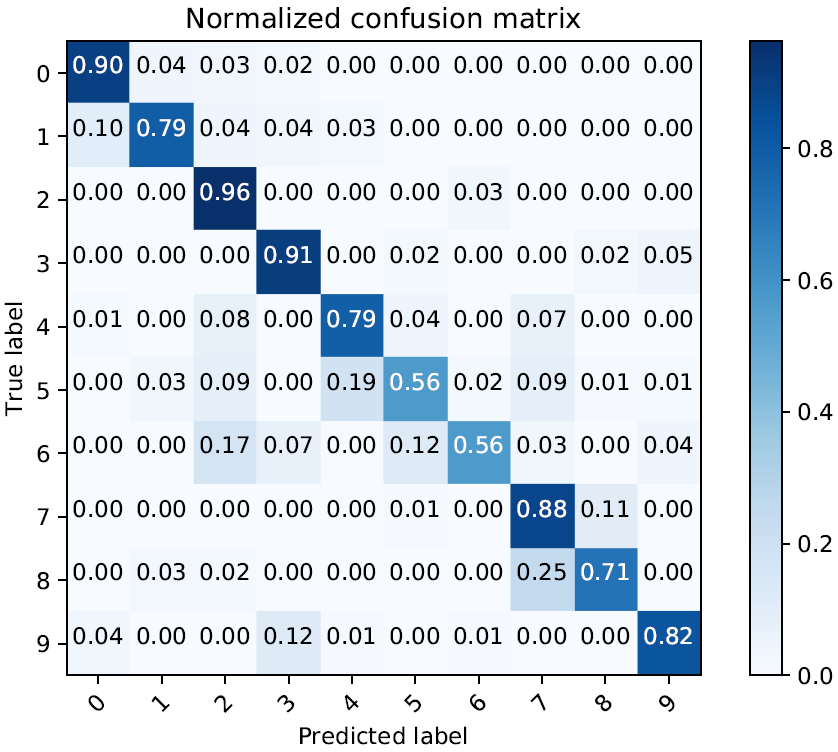}
\end{center}
   \caption{Deep network recognition accuracy and confusion matrix for the 10 body postures.}
\label{fig:confusion}
\end{figure}

Figure~\ref{fig:posture_speckle} shows the representative speckle patterns simulated from the different postures. Similar to the case of the MNIST hand-written digits, the speckle patterns from the same category of postures had some meaningful invariant features which could be used for recognition. Those speckle images were trained and tested using the ResNet-18 network. The training was performed on 11 out of 12 human subjects and tested on the remaining one human subject. This was repeated 12 times and the recognition rate was taken as the average of these 12 experiments. We achieved an average recognition accuracy of 78.18\%, which is higher than the three-way human body posture classification result reported in~\cite{Satat2}. The recognition rate could be further improved by increasing the number and diversity of human subjects. The confusion matrix of the 10 different postures (labels 0-9) is shown in Figure~\ref{fig:confusion}. These results demonstrate the potential of our approach to recognize complex objects.

\section{Conclusions}

We demonstrated direct object recognition without line-of-sight. Direct object recognition does not require forming a clear image, thereby greatly reduces the complexity of the system hardware. Such recognition is enabled by optical coherence, which is provided by illuminating an object with a laser. The resulting speckle pattern from the scattered light is analyzed using a deep-learning algorithm. 

To verify the feasibility and performance of this new approach, we quantitatively evaluated NLOS recognition based on simulations and experiments with MNIST hand-written digits and human body postures. The results show that the recognition accuracy can be well above 90\% for the MNIST data in various scenarios, and 78.18\% for the human body posture data. That the final recognition result was highly dependent on the deep-learning algorithm indicates that there are many opportunities to further improve the recognition accuracy with additional optimization of the deep network model, especially for objects with complex features. 

In addition to the advantages of this new approach, there are limitations which afford opportunities for future work. For example, noises (\eg vibration) in the experiment could cause drifting of the speckle patterns, which would reduce the recognition accuracy. Moreover, the light intensity could be greatly reduced after it diffusively reflects from the wall, resulting in a much lower signal-to-noise ratio and potentially lower recognition accuracy. A filter could be used to reject ambient light. To increase working distance and area, a laser with a diverging beam could be used, such as edge-emitting diode laser. More robust hardware and deep-learning algorithm will be needed to further improve the performance. 

In brief, this NLOS object recognition using coherent illumination opens a new gate to situations where direct line-of-sight is blocked, and provides plenty of opportunities for research and application.

{\small
\bibliographystyle{ieee_fullname}
\bibliography{cvpr_final_2019}
}

\end{document}